\documentclass{article}
\usepackage[T1]{fontenc} 
\usepackage[utf8]{inputenc} 
\usepackage{ismir,amsmath,cite,url}
\usepackage{graphicx}
\usepackage{color}
\usepackage{xcolor}
\usepackage{multirow}

\usepackage{booktabs}

\usepackage{amsmath}
\usepackage{amssymb}
\usepackage{mathtools}
\usepackage{amsthm}

\usepackage{algorithm}
\usepackage{algorithmic}

\usepackage{lineno}

\newcommand*{\x}{\mathbf{x}}
\newcommand*{\z}{\mathbf{z}}
\newcommand*{\R}{\mathbb{R}}
\newcommand*{\Z}{\mathcal{Z}}
\newcommand*{\eps}{\boldsymbol{\epsilon}}
\newcommand*{\m}{\mathbf{m}}

\newcommand{\STAB}[1]{\begin{tabular}{@{}c@{}}#1\end{tabular}}

\title{Unsupervised Composable Representations for Audio}

\oneauthor
{Giovanni Bindi, Philippe Esling} {Institut de Recherche et Coordination Acoustique-Musique (IRCAM) \\ CNRS UMR 9912, Sorbonne Université \\ {\tt \{bindi, esling\}@ircam.fr}}

\def\authorname{F. Author, S. Author, and T. Author}

\begin{document}

\maketitle
\begin{abstract}
Current generative models are able to generate high-quality artefacts but have been shown to struggle with compositional reasoning, which can be defined as the ability to generate complex structures from simpler elements. In this paper, we focus on the problem of compositional representation learning for music data, specifically targeting the fully-unsupervised setting. We propose a simple and extensible framework that leverages an explicit compositional inductive bias, defined by a flexible auto-encoding objective that can leverage any of the current state-of-art generative models. We demonstrate that our framework, used with diffusion models, naturally addresses the task of unsupervised audio source separation, showing that our model is able to perform high-quality separation. Our findings reveal that our proposal achieves comparable or superior performance with respect to other blind source separation methods and, furthermore, it even surpasses current state-of-art supervised baselines on signal-to-interference ratio metrics. Additionally, by learning an a-posteriori masking diffusion model in the space of composable representations, we achieve a system capable of seamlessly performing unsupervised source separation, unconditional generation, and variation generation. Finally, as our proposal works in the latent space of pre-trained neural audio codecs, it also provides a lower computational cost with respect to other neural baselines.
\end{abstract}
\section{Introduction}\label{sec:introduction}

Generative models recently became one of the most important topic in machine learning research. Their goal is to learn the underlying probability distribution of a given dataset in order to accomplish a variety of downstream tasks, such as sampling or density estimation. These models, relying on deep neural networks as their core architecture, have demonstrated unprecedented capabilities in capturing intricate patterns and generating complex and realistic data \cite{bondtaylor}. Although these systems are able to generate impressive results that go beyond the replication of training data, some doubts have recently been raised about their actual reasoning and extrapolation abilities  \cite{srivastava2023beyond, yuksekgonul2023when}. Notably, a critical question remains on their capacity to perform \textit{compositional reasoning}. The principle of compositionality states that the meaning of a complex expression is dependent on the meanings of its individual components and the rules employed to combine them \cite{comp2010def, janssen2012comp}. This concept also plays a significant role in machine learning \cite{lake2016building}, with a particular emphasis in the fields of NLP and vision. Indeed, compositionality holds a strong significance in the \textit{interpretability} of machine learning algorithms \cite{jesse2020compos}, ultimately providing a better understanding of the behaviour of such complex systems. In line with recent studies on compositional inductive biases \cite{hinton2021, lake2023baroni}, taking a compositional approach would allow to build better representation learning and more effective generative models, but research on compositional learning for audio is still lacking.

In this work, we specifically focus on the problem of compositional representation learning for audio and propose a generic and simple framework that explicitly targets the learning of composable representations in a fully unsupervised way. Our idea is to learn a set of low-dimensional latent variables that encode semantic information which are then used by a generative model to reconstruct the input. While we build our approach upon recent diffusion models, we highlight that our framework can be implemented with any state-of-the-art generative system. Therefore, our proposal effectively combines diffusion models and auto-encoders and represents, to the best of our knowledge, one of the first contributions that explicitly target the learning of unsupervised compositional semantic representations for audio. Although being intrinsically modality-agnostic, we show that our system can be used to perform \textit{unsupervised source separation} and we validate this claim by performing experiments on standard benchmarks, comparing against both unsupervised and supervised baselines. We show that our proposal outperforms all unsupervised methods, and even supervised methods on some metrics.
Moreover, as we are able to effectively perform latent source separation, we complement our decomposition system with a prior model that performs \textit{unconditional generation} and \textit{variation generation} \cite{mariani2024multisource}. Hence, our method is able to take an audio mixture as input, and generate several high-quality variations for one of the instrumental part only, effectively allowing to control regeneration of a source audio material in multi-instrument setups. Furthermore, we train a masking diffusion model in the latent space of composable representation and show that our framework is able to handle both decomposition and generation in an effective way without any supervision. We provide audio examples, additional experiments and source code on a supporting webpage\footnote{ \url{https://github.com/ismir-24-sub/unsupervised_compositional_representations}}

\section{Background}
\label{sec:back}

In this section, we review the fundamental components of our methodology. Hence, we briefly introduce the principles underlying diffusion models and a recent variation rooted in autoencoders, referred to as Diffusion Autoencoder \cite{preechakul2022diffusion}, which serves as the basis for our formulation.

\textbf{Notation.} Throughout this paper, we suppose a dataset $\mathcal{D} = \{\x_i\}_{i=1}^n$ of \textit{i.i.d.} data points $\x_i \in \R^d$ coming from an unknown distribution $p^*(\x)$. We denote $\theta \in \Theta \subseteq \R^p$, $\phi \in \Phi \subseteq \R^q$ and $\psi \in \Psi \subseteq \R^r$ as the set of parameters learned through back-propagation \cite{backprop}.

\subsection{Diffusion models}
Diffusion models (DMs) are a recent class of generative models that can synthesize high-quality samples by learning to reverse a stochastic process that gradually adds noise to the data. DMs have been successfully applied across diverse domains, including computer vision \cite{Karras2022edm}, natural language processing \cite{he-etal-2023-diffusionbert}, audio \cite{liu2023audioldm} and video generation \cite{ho2022video}. These applications span tasks such as unconditional and conditional generation, editing, super-resolution and inpainting, often yielding state of the art results. 

This model family has been introduced by \cite{sohl2015deep} and has its roots in statistical physics, but there now exist many derivations with different formalisms that generalise the original formulation. At their core, DMs are composed of a \textit{forward} and \textit{reverse} Markov chain that respectively adds and removes Gaussian noise from data. Recently, \cite{ho2020denoising} established a connection between DM and denoising score matching \cite{NEURIPS2019_3001ef25, vincent2011connection}, introducing simplifications to the original training objective and demonstrating strong experimental results. Intuitively, the authors propose to learn a function $\eps_\theta$ that takes a noise-corrupted version of the input and predicts the noise $\eps$ used to corrupt the data. Specifically, the \textit{forward} process gradually adds Gaussian noise to the data $\x \to \x_t$ according to an increasing noise variance schedule $\beta_1, \dots, \beta_T$, following the distribution
\begin{equation}
    q(\x_t | \x_{t-1}) = \mathcal{N}(\x_t; \sqrt{1 - \beta_t}\x_{t-1}, \beta_t \boldsymbol{I}),
\end{equation}
with $T \in \mathbb{N}$ and $t \in \{1, \dots, T\}$. Following the notation $\alpha_t = 1 - \beta_t$ and $\bar{\alpha}_t = \prod_{s=1}^t \alpha_s$, diffusion models approximate the \textit{reverse} process by learning a function $\eps_\theta : \R^d \times \R \to \R^d$ that predicts $\eps \sim \mathcal{N}(\eps, \mathbf{0}, \mathbf{I})$ by
\begin{equation}
    \min_{\theta \in \Theta} \quad \mathbb{E}_{t,\x_0,\eps} \big [ \| \eps_\theta(\sqrt{\bar{\alpha}_t} \x_0 + \sqrt{1 - \bar{\alpha}_t}\eps, t) - \eps \| \big ],
\end{equation}
with $\eps_\theta$ usually implemented as a U-Net \cite{ronneberger2015u} and the step $t\sim \mathcal{U}[0, T]$. 

\textbf{Deterministic diffusion.} More recently, \cite{song2021denoising} introduced Denoising Diffusion Implicit Models (DDIM), extending the diffusion formulation with non-Markovian modifications, thus enabling deterministic diffusion models and substantially increasing their sampling speed. They also established an equivalence between their objective function and the one from \cite{ho2020denoising}, highlighting the generality of their formulation. Finally, \cite{iadb} further generalized this approach and proposed Iterative $\alpha-$(de)Blending (IADB), simplifying the theory of DDIM while removing the constraint for the target distribution to be Gaussian. In fact, given a base distribution\footnote{For simplicity we assume $p_n(\x_0) = \mathcal{N}(\x_0; \boldsymbol{0}, \boldsymbol{I}).$} $p_n(\x_0)$, we corrupt the input data by linear interpolation $\x_\alpha = (1 - \alpha) \x_0 + \alpha \x$ with $\x_0 \sim p_n(\x_0)$ and learn a U-Net $\eps_\theta$ by optimizing, e.g.,
\begin{equation}
    \min_{\theta \in \Theta} \quad \mathbb{E}_{\alpha, \x, \x_0} \big [ \| \eps_\theta(\x_\alpha, \alpha) - \x \|_2^2 \big ],
\end{equation}
with $\alpha \sim \mathcal{U}[0, 1]$. This is known as the $c$ variant of IADB, which is the closest formulation to DDIM. In our implementation, we instead use the $d$ variant of IADB, which has a slightly different formulation that we do not report for brevity. We experimented with both variants and did not find significant discrepancies in performances.

\textbf{Diffusion Autoencoders.} All the methods described in the preceding paragraph specifically target unconditional generation. However, in this work we are interested in conditional generation and, more specifically, in a conditional encoder-decoder architecture. For this reason, we build upon the recent work by \cite{preechakul2022diffusion} named Diffusion Autoencoder (DiffAE). The central concept in this approach involves employing a learnable encoder to discover high-level semantic information, while using a DM as the decoder to model the remaining stochastic variations. Therefore, the authors equip a DDIM model $\epsilon_\phi$ with a semantic encoder $E_\theta : \R^d \to \R^s$ with $s \ll d$ that is responsible for compressing the high-level \textit{semantic} information\footnote{In the domain of vision this could be the identity of a person or the type of objects represented in an image.} into a latent variable $\z \in \R^s$ as $\z = E_\theta(\x)$. The DDIM model is, therefore, conditioned on such semantic representation and trained to reconstruct the data via 
\begin{equation}
    \min_{\theta \in \Theta, \phi \in \Phi} \quad \mathbb{E}_{t,\x_0,\eps} \big [ \| \eps_\phi(\sqrt{\alpha} \x_0 + \sqrt{1 - \alpha}\eps, \z, t) - \eps \| \big ]
\end{equation}
with $\alpha = \prod_{s=1}^t (1 - \beta_s)$ and $\beta_i$ being the variance at the $i-$th step. Since the DiffAE represents the state of the art for encoder-decoder models based on diffusion, we build our compositional diffusion framework upon this formulation, which we describe in the following section. 

\section{Proposed approach}
\label{sec:approach}

\begin{figure*}[t]
    \centering
    \includegraphics[width=0.98\textwidth]{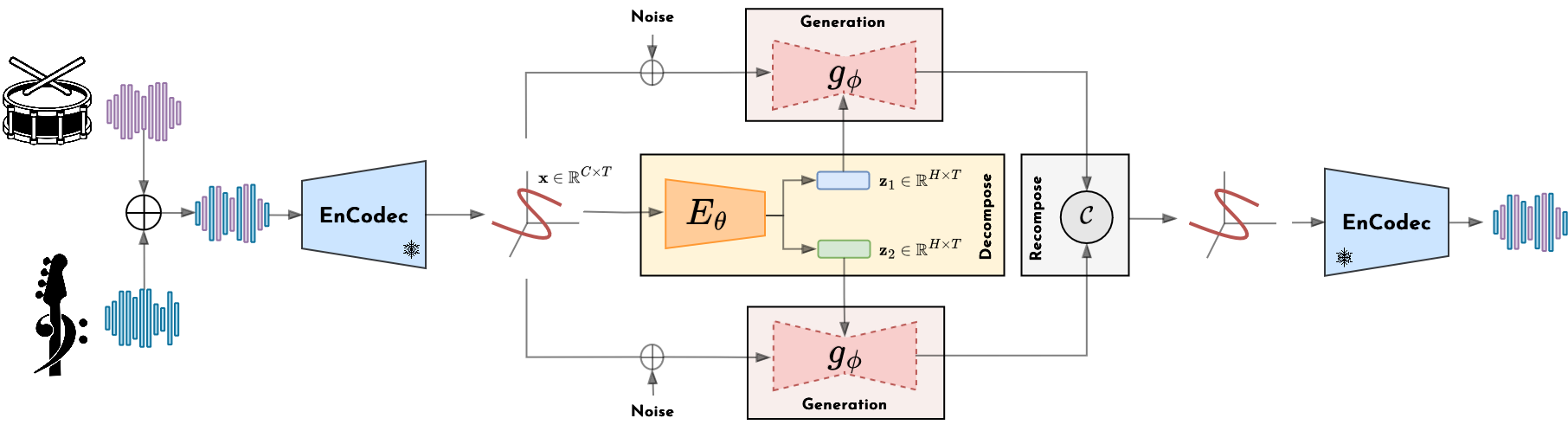}
    \caption{The overall architecture of our decomposition model. We first mix the sources, map the data $\x$ to the latent space through a frozen, pre-trained EnCodec model, and then decompose it into a set of latent variables (two shown here). These variables then condition a parameter-sharing diffusion model whose generation are then recomposed by an operator $C$.}
    \label{fig:model}
\end{figure*}

In compositional representation learning, we hypothesize that the information can be deconstructed into specific, identifiable parts that collectively makes up the whole input. In this work, we posit these parts to be distinct instruments in music but we highlight that this choice is uniquely dependent on the target application. Due to the lack of a widely-accepted description of compositional representations, we formulate a simple yet comprehensive definition that can subsequently be specialized to address particular cases \cite{andreas2018measuring, wiedemer2023compositional}. Specifically, we start from the assumption that observations $\x \in \R^d$ are realizations of an underlying latent variable model and that each concept is described by a corresponding latent $\z_i \in \Z_i$, where $i \in \{1, \dots, N\}$ with $N$ being the total number of possible entities that compose our data. Then, we define a compositional representation of $\x$ as
\begin{equation}
    \label{eq:comprep}
    \x = \mathcal{C}(\hat{\z}_1, \dots, \hat{\z}_N) = \mathcal{C}(f_1(\z_1), \dots, f_N(\z_N)),
\end{equation}
where $\mathcal{C} : \hat{\Z}_1 \times \hat{\Z}_2 \times \dots \hat{\Z}_N \to \R^d$ is a \textit{composition operator} and each $f_i : \Z_i \to \hat{\Z}_i$ is a \textit{processing function} that maps each latent variable to another \textit{intermediate} space. By being intentionally broad, this definition does not impose any strong specific constraints a priori, such as the requirement for each subspace to be identical or the algebraic structure of the latent space itself. Hence, to implement this model, we rather need to consider careful intentional design choices and inductive biases.
In this work, we constrain the intermediate space to be the data space itself, i.e. $\hat{\Z}_i = \R^d$ for all $i = 1, \dots, N$ and we focus on the learning of the latent variables and the processing functions. Finally, we set the composition operator to be a pre-defined function such as $mean$ or $max$ and leave its learning to further investigations.

\subsection{Decomposition}
\label{sec:diffdec}

In this section, we detail our proposed model, as depicted in Figure \ref{fig:model}. Globally, we follow an encoder-decoder paradigm, where we encode the data $\x \in \R^d$ into a set of latent representations $Z = \{\z_1, \dots, \z_N\}$, where $\z_i \in \Z \subseteq \R^h$ for each $i = 1, \dots, N$. This is done through an encoder network $E_\theta : \R^d \to \Z \times \dots \times \Z$ that maps the input $\x$ to the set of variables $Z$, i.e. $[\z_1, \dots, \z_N] = E_\theta(\x)$. Each latent variable is then decoded separately through a parameter-shared diffusion model, which implements the \textit{processing function} $f : \Z \to \R^d$ in Equation \ref{eq:comprep}, mapping the latents to the data space. Finally, we reconstruct the input data $\x$ through the application of a \textit{composition operator} $\mathcal{C}$ and train the system end-to-end through a vanilla iterative $\alpha-$(de)Blending (IADB) loss. Specifically, we learn a U-Net network $g_\phi : \R^d \times \R \times \R^h \to \R^d$ and a semantic encoder $E_\theta$ via the following objective
\begin{equation}
    \label{eqn:obj}
    \min_{\theta \in \Theta, \phi \in \Phi} \quad \mathbb{E}_{\alpha, \x, \x_0} \big [ \| \hat{g}_\phi(\x_\alpha, \alpha) - \x \|_2^2 \big ],
\end{equation}
with $\alpha \sim \mathcal{U}[0, 1]$, $\x_0 \sim \mathcal{N}(\x_0; \boldsymbol{0}, \boldsymbol{I})$ and
\begin{equation}
    \hat{g}_\phi(\x_\alpha, \alpha) = \mathcal{C}(g_\phi(\x_\alpha, \alpha, \z_1), \dots, g_\phi(\x_\alpha, \alpha, \z_N)),
\end{equation}
with $\x_\alpha = (1 - \alpha) \x_0 + \alpha \x$ and $[\z_1, \dots, \z_N] = E_\theta(\x)$.
We chose the IADB paradigm due to its simplicity in implementation and intuitive nature, requiring minimal hyper-parameter tuning.

At inference time, we reconstruct the input by progressively denoising an initial random sample coming from the prior distribution, conditioned on the components obtained through the semantic encoder.

\textbf{A note on complexity.} We found that using a single diffusion model proves effective instead of training $N$ separate models for $N$ latent variables. Consequently, we opt for training a parameter-sharing neural network $g_\phi$. Nonetheless, the computational complexity of our framework is therefore $N$ times that of a single DiffAE.

\subsection{Recomposition}

\label{sec:prior}

One of our primary objectives is to endow models with \textit{compositional generation}, a concept we define as the ability to generate novel data examples by coherently re-composing distinct parts extracted from separate origins. This definition aligns with numerous related studies that posit compositional generalization as an essential requirement to bridge the gap between human reasoning and computational learning systems \cite{fodor1988}. In this work, we allow for compositional generation by learning a prior model in the components' space. Specifically, once we have a well-trained decomposition model $D_{\theta,\phi} = (E_\theta, g_\phi)$ we learn a diffusion model in $\Z$ in order to obtain a full generative system. We define $\z = [\z_1, \dots, \z_N] = E_\theta(\x)$ and train a IADB model to recover $\z$ from a masked view $\tilde{\z}$. At training time, with probability $p_{mask}$, we mask each latent variable $\z_i$ with a mask $\m_i \in \{0, 1\}^{dim(\Z)}$ and optimize the diffusion model $\epsilon_\psi$ by solving
\begin{equation}
    \min_{\psi \in \Psi} \mathbb{E}_{\alpha, \z, \z_0, \mathbf{m}} [ \| \z - \epsilon_\psi(\z_\alpha, \alpha, \mathbf{m}) \|^2 ],
\end{equation}
where $\z_\alpha = \tilde{\z}_\alpha \odot \mathbf{m} + (1 - \mathbf{m}) \odot \z$ and $\tilde{\z}_\alpha = (1 - \alpha) \z_0 + \alpha \z$. Here, $\z_0 \sim \mathcal{N}(\z_0; \mathbf{0}, \mathbf{I})$ and $\tilde{\z}_\alpha$ denotes the $\alpha$-blended source $\z$. At each training iteration we randomly mask $\tilde{\z}_\alpha$ via $\m$ and train the diffusion model $\epsilon_\psi$ to recover the masked elements given the unmasked view $\z$. Our masking strategy allows for dropping each latent separately as well as all the latents simultaneously, effectively leading to a model that is able to perform both conditional and unconditional generation at the same time. In our application case, the conditional generation task reduces to the problem of generating variations. As our decomposition model proves to be effective in separating the stems of a given mixture, we obtain a system that is able to generate missing stems given the masked elements. Hence, this also addresses the accompaniment generation task. Algorithm \ref{alg:prior} resumes the training process of the prior model.

\begin{algorithm}[tb]
   \caption{Training prior model}
   \label{alg:prior}
\begin{algorithmic}
   \STATE {\bfseries Input:} dataset $\mathcal{D}$, U-Net $\epsilon_\psi$, pre-trained semantic encoder $E_\theta$, masking probability $p_{mask}$, learning rate $\gamma$.
   \WHILE{not converged}
   \FOR{$\x$ in $\mathcal{D}$}
   \STATE $\z = [\z_1, \dots, \z_N] = E_\theta(\x)$.
   \STATE Sample $\alpha \sim \mathcal{U}[0, 1]$ and $\z_0 \sim \mathcal{N}(\mathbf{0}, \mathbf{I})$.
   \STATE $\tilde{\z}_\alpha = (1 - \alpha) \z_0 + \alpha \z$
   \STATE Draw $\m \in \{0, 1\}^{dim(\Z) \times \dots \times dim(\Z)}$
   \STATE $\z_\alpha = \tilde{\z}_\alpha \odot \m + (1 - \m) \odot \z$
   \STATE $\mathcal{L}(\psi, \z, \alpha, \m) = \| \z - \epsilon_\psi(\z_\alpha, \alpha, \m) \|^2$
   \STATE Update $\psi \leftarrow \psi - \gamma \nabla_\psi \mathcal{L}(\psi, \z, \alpha, \m)$
   \ENDFOR
   \ENDWHILE
   \STATE \textbf{Return:} $\epsilon_\psi$
\end{algorithmic}
\end{algorithm}

\section{Experiments and Results}
\label{sec:exp}

This section provides an overview of the experiments aimed at assessing the performance of our proposal in both decomposition (section~\ref{sec:expdec}) and recomposition (section~\ref{sec:exprec}) scenarios. Prior to diving into the specifics of each experiment, we provide a brief overview of the shared elements across our experiments, including data, evaluation metrics, and neural network architectures.

\textbf{Data.} We rely on the Slakh2100 dataset \cite{manilow2019cutting}, a widely recognized benchmark in source separation, comprising 2100 tracks automatically mixed with separate stems. We selected this dataset because of its large-scale nature and the availability of ground truth separated tracks. Following recent approaches in generative models \cite{rombach2021highresolution, schneider2023mousai}, we rely on a pre-trained neural codec to map the audio data to an intermediate latent space, where we apply our approach. Specifically, we employ the EnCodec model \cite{defossez2022highfi}, a Vector Quantized-VAE (VQ-VAE) model \cite{denoord2017vq} that incorporates Residual Vector Quantization \cite{zeghidour2021soundstream} to achieve state-of-the-art performances in neural audio encoding. We take $24$ kHz mixtures from the Slakh2100 dataset, which we then feed to the pre-trained EnCodec model to extract the continuous representation obtained by decoding the discrete codes. EnCodec maps raw audio to latent trajectories with a sampling rate of $75$ Hz. Specifically, we take audio crops of approximately $7s$ $(6.82s)$, which are mapped via EnCodec to a latent code $\x \in \R^{128 \times 512}$. 

\textbf{Evaluation metrics.} Throughout this section, we report quantitative \textit{reconstruction} metrics in terms of both Mean Squared Error (MSE) and Multi-Scale Short-Time Fourier Transform (MS-STFT) \cite{yamamoto2020, caillon2021rave} for latent and audio data, respectively. We perform the MS-STFT evaluation using five STFT with window sizes $\{2048, 1024, 512, 256, 128\}$ following the implementation of \cite{caillon2021rave}. In order to evaluate the quality of the generated samples and the adherence to the training distribution, we also compute Fréchet Audio Distance (FAD) \cite{kilgour2019frechet, fadtk} scores. Specifically, we obtain the FAD scores via the \texttt{fadtk} library \cite{fadtk}, employing both the LAION-CLAP-Audio (LC-A) and LAION-CLAP-Music (LC-M) models \cite{laionclap2023}, as it was shown in \cite{fadtk} that these embedding models correlate well with perceptual tests measuring subjective quality of pop music. In assessing FAD scores, we utilize the complete test set of Slakh2100, while for MSE and MS-STFT values, we randomly select 512 samples of $7s$ ($\sim 1$ hour) from the same test set and report their mean and standard deviation. Finally, in order to provide the reader a reference value, we report in Table \ref{tab:encodecrec} the reconstruction metrics for the pre-trained EnCodec.

When assessing the effectiveness of \textit{source separation} models, we adhere to common practice by relying on the \texttt{museval} Python library \cite{SiSEC18} to compute standard separation metrics: Source-to-Interference Ratio (SIR), Source-to-Artifact Ratio (SAR), and Source-to-Distortion Ratio (SDR) \cite{roux2018sdr}. These metrics are widely accepted for evaluating source separation models, where SDR reflects sound quality, SIR indicates the presence of other sources, and SAR evaluates the presence of artifacts in a source. Specifically, following \cite{roux2018sdr} we compute their scale-invariant (SI) versions and, hence, provide our results in terms of SI-SDR, SI-SIR and SI-SAR. The values shown are expressed in terms of mean $\mu$ and standard deviation $\sigma$ computed on $512$ samples of $\sim 7s$ from the Slakh2100 test set. 

\begin{table}[]
    \centering
    \begin{tabular}{ccc}
    \toprule
     MS-STFT & FAD {\tiny (LC-A)} & FAD {\tiny (LC-M)} \\
    \midrule
     4.7 & 0.05 & 0.04 \\
    \bottomrule
    \end{tabular}
    \caption{EnCodec reconstruction quality, measured in terms of MS-STFT and FAD and computed following the procedure descried in section \ref{sec:exp}.}
    \label{tab:encodecrec}
\end{table}

\textbf{Architectures.} We use a standard U-Net \cite{ronneberger2015u} with 1D convolution and an encoder-decoder architecture with skip connections. Each processing unit is a ResNet block \cite{he2016deep} with group normalization \cite{wu2018group}. Following \cite{dhariwal2021diffusion}, we feed the noise level information through Positional Encoding \cite{vaswani2017attention}, conditioning each layer with the AdaGN mechanism. We also add multi-head self-attention \cite{vaswani2017attention} in the bottleneck layers of the U-Net. The semantic encoder mirrors the U-Net encoder block without the attention mechanism and maps the data $\x \in \R^{128 \times 512}$ to a set of variables $\z = [\z_1 \dots \z_i \dots \z_N]$ whose dimensionality is $\z_i \in \R^{1 \times 512}$. Finally, these univariate latent variables condition the U-Net via a simple concatenation, which proved to be a sufficiently effective conditioning mechanism for the model to converge. We use the same U-Net architecture for both the decomposition and recomposition diffusion models.

\subsection{Decomposition}
\label{sec:expdec}
In order to show the effectiveness of our decomposition method described in section \ref{sec:diffdec}, we perform multiple experiments on Slakh2100. Throughout this section, we fix the number of training epochs to $250$ and use the AdamW optimizer \cite{loshchilov2018decoupled} with a fixed learning rate of $10^{-4}$ as our optimization strategy. The U-Net and semantic encoder have $13$ and $8$ million trainable parameters, respectively. Finally, we use $100$ sampling steps at inference time.

First, we show in Table \ref{tab:sourcesep} that our model can be used to perform unsupervised latent source separation and compare it against several non-neural baselines \cite{rpca, lee1999learning, hpss2010, repet2012, 2dft}, as well as a recent study that explicitly targets neural latent blind source separation \cite{lass2023}. We also report the results obtained by Demucs \cite{defossez2020music}, which is the current top performing fully-supervised state-of-the-art method in audio source separation. As the only non-neural baseline, LASS, has been trained and evaluated on the Drums + Bass subset, we perform our analysis on this split and subsequently perform an ablation study over the other sources.

\begin{table}[!h]
  \centering
  \begin{tabular}{lccc}
    \toprule
     Model & SI-SDR ($\uparrow$) & SI-SIR ($\uparrow$) & SI-SAR ($\uparrow$) \\
    \midrule
     rPCA \cite{rpca} & -2.8 (4.8) & 5.2 (7.3) & \underline{5.6} (4.6) \\
     REPET \cite{repet2012} & -0.5 (4.8) & 6.8 (7.0) & 3.0 (5.2) \\
     FT2D \cite{2dft} & -0.2 (4.7) & 5.1 (7.0) & 3.1 (4.7) \\
     NMF \cite{lee1999learning}  & 1.4 (5.0) & 8.9 (7.6) & 2.9 (4.5) \\
     HPSS \cite{hpss2010} & 2.3 (4.8) & 9.9 (7.5) & 5.1 (4.6) \\
     \midrule
     LASS \cite{lass2023}  & -3.3 (10.8) & 17.7 (11.6) & -1.6 (11.2) \\
     Ours & \underline{5.5} (4.6) & \textbf{41.7} (9.3) & \underline{5.6} (4.6) \\
     \midrule
     Demucs \cite{defossez2020music}   & \textbf{11.9} (5.0) & \underline{37.6} (8.7) & \textbf{12.0} (5.0) \\
     \bottomrule
  \end{tabular}
    \caption{Blind source separation results for the \textit{Drums + Bass} subset. Our model is trained with the \textit{mean} composition operator. The results are expressed in dB as the mean (standard deviation) across $512$ elements randomly sampled from the test set of Slakh2100.}
    \label{tab:sourcesep}
\end{table} 

As we can see, our model outperforms the other baselines in terms of SI-SDR and SI-SIR and performs on par with respect to SI-SAR. Interestingly, our model outperforms the Demucs supervised baseline in terms of SI-SIR, which is usually interpreted as the amount of other sources that can be heard in a source estimate. In order to test LASS performances, we used their open source checkpoint which is trained on the Slakh2100 dataset, and followed their evaluation strategy. Unfortunately, we were not able to reproduce their results in terms of SDR but we found that their model performs well in terms of SI-SIR, which they did not measure in the original paper. Moreover, as LASS comprises training one transformer model per source, we found their inference phase to be more computationally demanding than ours. Finally, among non-neural baselines, we see that the HPSS model outperforms the others. This seems reasonable as HPSS is specifically built for separating percussive and harmonic sources and hence naturally fits this evaluation context.

Moreover, in order to show the robustness of our approach against different sources and number of latent variables, we train multiple models on different subset of the Slakh2100 dataset, namely \textit{Drums + Bass, Piano + Bass} and \textit{Drums + Bass + Piano}. The interested reader can refer to our supplementary material and listen to the separation results.

Subsequently, we show that our objective in Equation \ref{eqn:obj} is robust across different composition operators. We show that, for simple functions such as \textit{sum, min, max} and \textit{mean} our model is able to effectively converge and provide accurate reconstructions. Again, we provide this analysis by training our model on the \textit{Drums + Bass} subset of Slakh2100, fixing the number of components to $2$. We report quantitative results in terms of two reconstruction metrics, the Mean Squared Error (MSE) and Multi-Scale STFT distance (MS-STFT) in Table \ref{tab:recqual}. As we can see, \textit{sum} and \textit{mean} operators provided the best results, while \textit{min} and \textit{max} proved to be less effective. Nonetheless, the audio reconstruction quality measured in terms of MS-STFT provided reconstruction scores that are lower or comparable with respect to those obtained by evaluating EnCodec performances.

\begin{table}[]
    \centering
    \begin{tabular}{lcc}
    \toprule
    Operator & MSE ($\downarrow$) $\times 10^4$ & MS-STFT ($\downarrow$) \\
    \midrule
    \textit{Sum} & 1.87820 (0.13418) & 3.6 (0.1) \\
    \textit{Mean} & 1.87020 (0.13183) & 3.6 (0.1) \\
    \textit{Min} & 2.54182 (0.17714) & 4.5 (0.1) \\
    \textit{Max} & 2.43302 (0.17510) & 4.3 (0.1) \\
    \bottomrule
    \end{tabular}
    \caption{Reconstruction quality in latent space (MSE) and audio (MS-STFT) of our decomposition-recomposition model for different recomposition operators for the \textit{Drums + Bass} subset.}
    \label{tab:recqual}
\end{table}

\begin{table*}[]
\centering
\begin{tabular}{lcc|cc}
\toprule
  & \multicolumn{2}{c}{Original} & \multicolumn{2}{c}{Encoded} \\
       & FAD {\tiny (LC-A)} ($\downarrow$)     & FAD {\tiny (LC-M)} ($\downarrow$)    & FAD {\tiny (LC-A)} ($\downarrow$)       & FAD {\tiny (LC-M)} ($\downarrow$)       \\ \hline
Unconditional & 0.09         & 0.09        & 0.06           & 0.06           \\
$p_{mask} = 0.8$  & 0.12         & 0.11        & 0.08           & 0.07 \\
\midrule
Bass & 0.03 & 0.03 & 0.01 & 0.01 \\
Drums & 0.09 & 0.08 & 0.05 & 0.05 \\
\bottomrule
\end{tabular}
\caption{Audio quality of unconditional generations by our generative model. We demonstrate that we can jointly learn an unconditional and conditional model by showing that the FAD scores of $p_{mask} = 0.8$ are comparable to those of an unconditional latent diffusion model.}\label{tab:uncondquality}
\end{table*}

\begin{table}[!h]
  \centering
  \begin{tabular}{clcc}
    \toprule
     & \multicolumn{1}{c}{Type} & \multicolumn{1}{c}{MSE $\times 10^3$}  & \multicolumn{1}{c}{MS-STFT} \\
    \midrule
    \multirow{2}{*}{\STAB{\rotatebox[origin=c]{90}{\textbf{Real}}}}
     & Drums & 2.3259 (0.1287) & 13.6 (0.4) \\
     & Bass & 1.4393 (0.0874) & 9.38 (0.2) \\
     \midrule
     \multirow{2}{*}{\STAB{\rotatebox[origin=c]{90}{\textbf{Rand}}}}
     & Drums & 4.8170 (0.1136) & 20.5 (0.6) \\
     & Bass & 4.8814 (0.1157) & 21.7 (0.7) \\
     \bottomrule
  \end{tabular}
    \caption{Diversity of variations generated by our prior model, measured via the MSE and MS-STFT distances against ground truth and random components.}
    \label{tab:diversity}
\end{table} 

\subsection{Recomposition} 
\label{sec:exprec}
As detailed in section \ref{sec:prior}, once we are able to decompose our data into a set of composable representations we can then learn a prior model for generation from this new space. Since our decomposition model is able to compress meaningful information through the semantic encoder, we can learn a second latent diffusion model on this compressed representation to obtain a full generative model able to both decompose and generate data.

Here, we validate our claims by training a masked diffusion model for the \textit{Drums + Bass} split of the Slakh2100 dataset. In Table \ref{tab:uncondquality}, we show that our model can indeed produce good-quality unconditional generations by comparing it against a fully unconditional model. We measure the generation quality in terms of FAD scores computed against both the original as well as the encoded test data. Here, by original data we mean the audio coming from the test split of Slakh2100, while the encoded data represents the same elements reconstructed with our decomposition algorithm. As we train on the representations obtained through the semantic encoder, the natural benchmark for the unconditional generation is given by the reconstructions that we can obtain through our decomposition model, which represents the bottleneck in terms of quality. Nonetheless, we show that the FAD scores do not drop substantially when comparing against the original audio, showing that we can indeed achieve a good generation quality.
In the same table, we report the partial generation FAD scores. Instead of generating both components unconditionally, we generate the Bass (Drums) given the Drums (Bass), and measure the FAD against the original and the encoded test data, as done for the unconditional case. Given the presence of a ground-truth element, the FAD scores are lower, which is to be expected. Specifically, we can see that the drums generation is a more complex task with respect to the bass generation, as the model needs to synthesize more elements such as the kick, snare and hi-hats, matching the timing of a given bassline.

Lastly, as we strive for high-quality generations, we also aim to enhance diversity within our generations.
Table \ref{tab:diversity} shows the diversity scores for partial generations obtained with our model. We measure diversity in terms of MSE and MS-STFT scores computed, respectively, in the latent and audio space. We compare our partial generations against real and random components, in order to provide the lower and upper bound for generation diversity. Specifically, given the Drums (Bass) we generate the Bass (Drums) and we compute both MSE and MS-STFT scores against the ground truth (Real) and random elements (Rand) coming from the test set of Slakh2100. From the values reported in Table \ref{tab:diversity}, we can deduce that our model produces meaningful variations. We invite the interested readers to listen to our results on our support website.

\section{Discussion and Further Works}
\label{sec:discussion}

While our model proves to be effective for compositional representation learning, it still has shortcomings. Here, we briefly list the weaknesses of our proposal and highlight potential avenues for future investigations. 

\textbf{Factors of convergence.} In this paper, we used EnCodec which already provides some disentanglement and acts as a sort of initialization strategy for our method. We argue that this property, jointly with the low dimensionality of the latent space enforced by our encoder leads our decomposition model to converge efficiently, not requiring further inductive biases towards source separation.

\textbf{Limitations.} First, there is no theoretical guarantee that the learned latent variables are bound to encode meaningful information. Exploring more refined approaches, as proposed by \cite{wang23infodiff}, could be interesting in order to incorporate a more principled method for learning disentangled latent representations. Furthermore, we observed that the dimensionality of the latent space significantly influences the representation content. A larger dimensionality allows the model to encode all the information in each latent, hindering the learning of distinct factors. Conversely, a smaller dimensionality may lead to under-performance, preventing the model to correctly converge. It could be interesting to investigate strategies such as Information Bottleneck \cite{tishby2000information} to introduce a mechanism to explicitly trade off expressivity with compression. Finally, using more complex functions as well as learnable operators is an interesting research direction for studying the interpretability of learned representations.

\section{Conclusions}
\label{sec:conc}

In this work, we focus on the problem of learning unsupervised compositional representations for audio. We build upon recent state-of-the-art diffusion generative models to design an encoder-decoder framework with an explicit inductive bias towards compositionality. We validate our approach on audio data, showing that our method can be used to perform latent source separation. Despite the theoretical shortcomings, we believe that our proposal can serve as a useful framework for conducting research on the topics of unsupervised compositional representation learning.

\bibliography{ISMIRtemplate}

\end{document}